\def\year{2019}\relax
\documentclass[conference,letterpaper]{IEEEtran}
\usepackage{cite}
\usepackage{adjustbox} 
\usepackage{amsmath,amssymb,amsfonts}
\usepackage{algorithmic}
\usepackage{graphicx}
\usepackage{textcomp}
\usepackage{xcolor}

\def\BibTeX{{\rm B\kern-.05em{\sc i\kern-.025em b}\kern-.08em
    T\kern-.1667em\lower.7ex\hbox{E}\kern-.125emX}}

\begin{document}
%
\title{On the Performance of Differential Evolution for Hyperparameter Tuning}
\author{
    \IEEEauthorblockN{Mischa Schmidt, Shahd Safarani,  Julia Gastinger, Tobias Jacobs, S\'{e}bastien Nicolas, Anett Sch\"ulke }
    \IEEEauthorblockA{NEC Laboratories Europe GmbH, Kurf\"ursten-Anlage 36, 69115 Heidelberg, Germany
    \\\{FirstName.LastName\}@neclab.eu}    
}
\maketitle
\begin{abstract}
Automated hyperparameter tuning aspires to facilitate the application of machine learning for non-experts. In the literature,  different optimization approaches are applied for that purpose. This paper investigates the performance of Differential Evolution for tuning hyperparameters of supervised learning algorithms for classification tasks. This empirical study involves a range of different machine learning algorithms and datasets with various characteristics to compare the performance of Differential Evolution with Sequential Model-based Algorithm Configuration (SMAC), a reference Bayesian Optimization approach. The results indicate that Differential Evolution outperforms SMAC for most datasets when tuning a given machine learning algorithm - particularly when breaking ties in a first-to-report fashion. Only for the tightest of computational budgets SMAC performs better. On small datasets, Differential Evolution outperforms SMAC by 19\% (37\% after tie-breaking). In a second experiment across a range of representative datasets taken from the literature, Differential Evolution scores 15\% (23\% after tie-breaking) more wins than SMAC. 
\end{abstract}

\section{Introduction}
When applying machine learning, several high-level decisions have to be taken: a learning algorithm needs to be selected (denoted \textit{base learner}), and different data preprocessing and feature selection techniques may be applied. Each comes with a set of parameters to be tuned - the \textit{hyperparameters}. Automated machine learning (AutoML) addresses the automation of selecting base learners and preprocessors as well as tuning the associated hyperparameters. That allows non-experts to leverage machine learning. For example, \cite{luo2016review} surveys AutoML approaches from the perspective of the biomedical application domain to guide ``layman users''. 
Furthermore, AutoML makes the process of applying machine learning more efficient, e.g., by using automation to lower the workload of expert data scientists. Additionally, AutoML provides a structured approach to identify well-performing combinations of base learner configurations, typically outperforming those tuned manually or by a grid search heuristic.



This work focuses on the AutoML step of tuning the hyperparameters of individual base learners. 
In most of the recent works on AutoML, model-based approaches are employed, such as Bayesian Optimization \cite{snoek2012practical} (see \cite{HutHooLey11}, \cite{feurer2015efficient}) or, more recently, Probabilistic Matrix Factorization \cite{NIPS2018_7595}. In addition to hyperparameter tuning they address features such as identifying best machine learning pipelines. Furthermore, \cite{feurer2015efficient} and \cite{NIPS2018_7595} also draw on information about machine learning tasks solved in the past. We are particularly interested in whether Bayesian Optimization is better suited for hyperparameter tuning than, e.g., nature-inspired black-box optimization heuristics. We motivate this by the observation that the work presented in \cite{NIPS2018_7595} discretizes continuous hyperparameters, which effectively turns them into categorical parameters and thereby gives up on the ambition to find the best performing hyperparameter configuration. Yet, this approach outperforms \cite{feurer2015efficient}, a widely used Bayesian Optimization approach, which allows for continuous and categorical hyperparameters. As \cite{NIPS2018_7595} (discretizing the space of machine learning pipeline configurations and thus being suboptimal) outperforms \cite{feurer2015efficient} (capable of dealing with fixed and categorical parameters), it calls into question whether the popular model-based approaches - in particular variants of Bayesian Optimization - appropriately reflect the base learner performance. While a variety of AutoML approaches described in literature are relying on a variety of methods ranging over Bayesian Optimization {\cite{snoek2012practical,feurer2015efficient}}, Probabilistic Matrix Factorization~\cite{NIPS2018_7595}, and Evolutionary Algorithms~\cite{olson2016tpot}, we are not aware of any direct comparison of these optimization approaches dedicated to hyperparameter tuning. 
Specifically, this paper investigates the performance of Differential Evolution \cite{Storn1997}, a representative evolutionary algorithm (i.e., a gradient-free black-box method), relative to Sequential Model-based Algorithm Configuration (SMAC) \cite{HutHooLey11} as a reference for Bayesian Optimization. The paper focuses exclusively on cold start situations to gain insights into each method's ability to tune hyperparameters.

\section{Related work}
Auto-sklearn \cite{feurer2015efficient} is probably the most prominent example of applying Bayesian Optimization (through the use of SMAC \cite{HutHooLey11}) for the automated configuration of machine learning pipelines. It supports reusing knowledge about well performing hyperparameter configurations when a given base learner is tested on similar datasets (denoted \textit{warm starting} or \textit{meta-learning}), ensembling, and data preprocessing. For base learner implementations \cite{feurer2015efficient} leverages scikit-learn \cite{pedregosa2011scikit}. \cite{feurer2015efficient} studies individual base learners' performances on specific datasets, but does not focus exclusively on tuning hyperparameters in its experiments. This work takes inspiration from \cite{feurer2015efficient} for Experiment 1 (base learner selection) and 2 (dataset and base learner selection).

A recent approach to modeling base learner performance applies a concept from recommender systems denoted \emph{Probabilistic Matrix Factorization}~\cite{NIPS2018_7595}. It discretizes the space of machine learning pipeline configurations and - typical for recommender systems - establishes a matrix of tens of thousands of machine learning pipeline configurations' performances on hundreds of datasets. Factorizing this matrix allows estimating the performance of yet-to-be-tested pipeline-dataset combinations. On a hold out set of datasets \cite{NIPS2018_7595} outperforms~\cite{feurer2015efficient}. We do not include Probabilistic Matrix Factorization in this work as recommender systems only work well in settings with previously collected correlation data, which is at odds with our focus on cold start hyperparameter tuning settings.

TPOT \cite{olson2016tpot} uses Genetic Programming, an evolutionary algorithm, to optimize machine learning pipelines (i.e., data preprocessing, algorithm selection, and parameter tuning) for classification tasks, achieving competitive results. To keep the pipelines' lengths within reasonable limits, it uses a fitness function that balances pipeline length with prediction performance. Similar to our work, TPOT relies on the DEAP framework~\cite{DEAP} but uses a different evolutionary algorithm. While hyperparameter tuning is an aspect of TPOT, \cite{olson2016tpot} does not attempt to isolate tuning performance and compare against other hyperparameter tuning approaches.

BOHB \cite{falkner2018bohb} is an efficient combination of Bayesian Optimization with Hyperband \cite{li2017hyperband}. In each BOHB iteration, a multi-armed bandit (Hyperband) determines the number of hyperparameter configurations to evaluate and the associated computational budget. This way, configurations that are likely to perform poorly are stopped early. Consequently, promising configurations receive more computing resources. The identification of configurations at the beginning of each iteration relies on Bayesian Optimization. Instead of identifying ill-performing configurations early on, our work focuses on the hyperparameter tuning aspect. In particular, we study empirically whether alternative optimization heuristics such as evolutionary algorithms can outperform the widely used model-based hyperparameter tuning approaches. 

This work differs from the referenced articles in that it attempts to isolate the hyperparameter tuning methods' performances, e.g., by limiting CPU resources (a single CPU core) and tight computational budgets (smaller time frames than in \cite{feurer2015efficient} and \cite{NIPS2018_7595}). These tightly limited resources are vital to identifying the algorithmic advantages and drawbacks of different optimization approaches for hyperparameter tuning. Different to, e.g., \cite{NIPS2018_7595} we do not limit the invocation of individual hyperparameter configurations. That penalizes ill-performing but computationally expensive parameter choices. To the best of our knowledge, such scenarios have not been studied in the related literature.

\section{Methods}

\subsection{Hyperparameter Tuning Definition}
Given a validation set, the performance of a trained machine learning algorithm is a function $f: \mathcal{X} \rightarrow \mathbb{R}$ of their continuous and categorical hyperparameters $x \in \mathcal{X}$ \cite{falkner2018bohb}. Therefore, hyperparameter tuning corresponds to finding the best performing algorithm configuration, i.e., $\arg \min_{x \in \mathcal{X}} f(x)$ (if using an error metric), or  $\arg \max_{x \in \mathcal{X}} f(x)$ (if using an accuracy metric, e.g., for classification tasks).

\subsection{Hyperparameter Tuning Methods}
\subsubsection{Evolutionary Algorithms}
This work applies evolutionary algorithms for hyperparameter tuning.  These are a subset of the vast range of nature-inspired meta-heuristics for optimization. Typically, candidate solutions $x \in \mathcal{X}$ are managed in a \textit{population}. New solution candidates evolve from the current population by using algorithmic operations that are inspired by the concepts of biological evolution: \textit{reproduction}, \textit{mutation}, \textit{recombination}, and \textit{selection}. Commonly, a problem-specific \textit{fitness function} (e.g., the performance $f(x)$) determines the quality of the solutions. Iterative application of the operations on the population results in its evolution. Evolutionary algorithms differ in how the algorithmic operations corresponding to the biological concepts are implemented. 


This work uses Differential Evolution \cite{Storn1997}, a well-known and well-performing direction-based meta-heuristic supporting real-valued as well as categorical hyperparameters \cite{5601760}\cite{ALDABBAGH2018284}, as a representative evolutionary algorithm. Unlike traditional evolutionary algorithms, Differential Evolution perturbs the current-generation population members with the scaled differences of randomly selected and distinct population members, and therefore no separate probability distribution has to be used for generating the offspring's genome \cite{5601760}.  Differential Evolution has only a few tunable parameters as indicated below, hence this work does not require extensive tuning of these parameters for the experimentation. Notably, variants of Differential Evolution produced state of the art results on a range of benchmark functions and real-world applications~\cite{ALDABBAGH2018284}. While several advanced versions of Differential Evolution exist \cite{5601760}\cite{ALDABBAGH2018284}, this work focuses on the basic variant using the implementation provided by the python framework DEAP~\cite{DEAP}. Should results indicate performance competitive to the standard Bayesian Optimization-based hyperparameter tuning approach, it will be a promising direction for future work to identify which of the variants is best suited for the hyperparameter tuning problem. 

Differential Evolution is derivative-free and operates on a population of fixed size $N$. Each population member of dimensionality $d$ represents a potential solution to the optimization problem. In this paper, $d$ equals the base learner's number of tunable hyperparameters  ($h$). We choose the population size depending on the base learner's number of hyperparameters ($h$): $N = n h$ ($= n d$ in this work), where $n$ is a configurable parameter. For generating new candidate solutions, Differential Evolution selects four population members to operate on: the \textit{target} with which the offspring will compete, and three other randomly chosen input population members. 
First, Differential Evolution creates a mutant by modifying one of the three input members. It modifies the mutant's values along each dimension by a fraction $DE_f$ of an application specific distance metric between both remaining input members. Then, the crossover operation evolves the mutant into the offspring: each of the mutant's dimensions' values may be replaced with probability $DE_{cr}$ by the target's corresponding value. The newly created offspring competes with the target to decide whether it replaces the target as a population member or whether it is discarded. \cite{eas-intro} provides detailed information on Differential Evolution and its operations.  

\subsubsection{Model-based}
Bayesian Optimization is very popular for hyperparameter tuning. In each iteration $i$, it uses a probabilistic model $p_{i-1}(f | \mathcal{D})$ to model the objective function $f$ based on observed data points $\mathcal{D} = \{ (x_0 , y_0 ), \dots , (x_{i-1} , y_{i-1} ) \}$. An acquisition function $a: \mathcal{X} \rightarrow \mathbb{R}$ based on the current $p_i(f | D)$ identifies the next $x_i$ - typically by $x_{i} = \arg \max_{x \in \mathcal{X}} a(x)$. The identified $x_i$ represents the next hyperparameter configuration to evaluate (i.e., to train and test) the machine learning algorithm with, i.e., $y_i = f(x_i)$. Note that observations of~$y_i$ may be noisy, e.g. due to stochasticity of the learning algorithm. After evaluation, Bayesian Optimization updates its model  $p_{i}(f | \mathcal{D} \cup (x_i , y_i ))$. A common acquisition function is the \textit{Expected Improvement} criterion  \cite{snoek2012practical}. 

For image classification \cite{snoek2012practical} obtained state-of-the-art performance for tuning convolutional neural networks, utilizing Gaussian Processes to model $p_i(f | D)$. Other approaches employing, e.g., the Tree Parzen Estimator technique \cite{bergstra2011algorithms} do not suffer some of the drawbacks of Gaussian Processes (e.g., cubic computational complexity in the number of samples). Similarly, the SMAC library \cite{HutHooLey11} uses a tree-based approach to Bayesian Optimization. \cite{falkner2018bohb} provides additional details. 

This paper relies on SMAC as a representative of Bayesian Optimization as it is a core constituent of the widely used auto-sklearn library \cite{feurer2015efficient}. SMAC ``iterates between building a model and gathering additional data'' \cite{HutHooLey11}. Given a base learner and a dataset, SMAC builds a model based on past performance observations of already trained and tested hyperparameter configurations. It optimizes the Expected Improvement criterion to identify the next promising hyperparameter configuration. That causes SMAC to search regions in the hyperparameter space where its model exhibits high uncertainty. After trying the identified configuration with the base learner, SMAC updates the model again.

\subsection{Experimental Setup and Evaluation}
This work focuses on cold start situations to isolate the aspect of tuning a given base learner's hyperparameters. Consequently, the experiments do not cover other beneficial aspects such as meta-learning, ensembling, and preprocessing steps. We denote the application of a hyperparameter tuning method to a pre-selected base learner on a specific dataset as an \textit{experiment run}. To study the hyperparameter tuning performances of Differential Evolution and SMAC, we assign equal computational resources (a single CPU core, fixed wall-clock time budget) to each experiment run. For assessing Differential Evolution's ability to tune hyperparameters relative to SMAC, we compare both methods' performance per experiment run by applying relative \textit{ranking}. Similar to \cite{feurer2015efficient}, we account for class occurrence imbalances using the \textit{balanced classification error} metric, i.e., the average class-wise classification error. The run with the better evaluation result (based on five-fold cross-validation) counts as a win for the corresponding tuning method. In each experiment, we create five data folds using the scikit-learn library \cite{pedregosa2011scikit}. The folds serve as input to both tuning methods' experiment runs. To break ties of reported results, we award the method requiring less wall-clock time to reach the best result with a win. This work experiments with six base learners provided by \cite{pedregosa2011scikit}: k-Nearest Neighbors (kNN), linear and kernel SVM, AdaBoost, Random Forest, and Multi-Layer Perceptron (MLP). 

This study documents two sets of classification experiments. In both, only the overall experiment run wall-clock time budget limits execution time, i.e., we do not limit the time taken for each base learner invocation (training and testing). 
We select the Differential Evolution hyperparameters as $n=10$, $DE_f=0.5$, and $DE_{cr}=0.25$ after a brief hyperparameter sweep. 

\textbf{Experiment 1}. This experiment executes a single experiment run of both tuning methods for each of the base learners for 49 small datasets (less than 10,000 samples) as identified by \cite{NIPS2018_7595} \footnote{www.OpenML.org datasets: \{23, 30, 36,  48, 285, 679, 683, 722, 732, 741, 752, 770, 773, 795, 799, 812, 821, 859, 862, 873, 894, 906, 908, 911, 912, 913, 932, 943, 971, 976, 995, 1020, 1038, 1071, 1100, 1115, 1126, 1151, 1154, 1164, 1412, 1452, 1471, 1488, 1500, 1535, 1600, 4135, 40475\}}. As these datasets are small, we assign a strict wall-clock time budget of one hour to each experiment run (one-third of the approximate time budget of \cite{NIPS2018_7595}). 

\textbf{Experiment 2}. The second experiment leverages the efforts of \cite{feurer2015efficient}, which identified representative datasets from a range of application domains to demonstrate the robustness and general applicability of AutoML. For that, \cite{feurer2015efficient} clustered 140 openly available binary and multiclass classification datasets covering a diverse range of applications, such as text classification, digit and letter recognition, gene sequence and RNA classification, and cancer detection in tissue samples based on the datasets' meta-features into 13 clusters. A representative dataset represents each cluster. Of these 13 representative datasets, we select ten\footnote{www.OpenML.org datasets: \{46, 184, 293, 389, 554, 772, 917, 1049, 1120, 1128\}}  not requiring preprocessing such as imputation of missing data and apply all six base learners to each. All experiment runs receive a time budget of 12 hours (half of the budget in \cite{feurer2015efficient}). We repeat this experiment for each base learner and dataset five times per tuning method. Per repetition, five data folds are created and presented to both tuning methods so that they face the same learning challenge.

\textbf{Generalization of results}. In total, the experiments cover 59 openly available datasets from www.openml.org as identified in \cite{feurer2015efficient} and \cite{NIPS2018_7595}. To help generalize the findings in this study, we treat each pairwise experiment run as a Bernoulli random variable and apply statistical bootstrapping (10,000 bootstrap samples, 95\% confidence level) to infer the confidence intervals for the probability of Differential Evolution beating SMAC. As the dataset selection criteria differ between Experiment 1 and 2, we will discuss the experiments' results separately.



\section{Experiment Results}

\subsection{Experiment 1}
Table \ref{tab:experiment1} documents the results of running both tuning algorithms for individual base learners on the 49 small datasets, each with a one hour wall-clock budget on a single CPU core. 

When considering only the maximum mean five-fold balanced error to rank the tuning methods, Differential Evolution (denoted \textit{DE} in Table~\ref{tab:experiment1} and~\ref{tab:budget_100}) scores 19.4\% more wins than SMAC (129 wins to 108). 19.4\% of the experiment runs result in a tie where both tuners achieve the same error. Breaking these ties (based on which tuning method reached its best result first) shows the benefits of Differential Evolution even clearer. It scores 37.1\% more wins than SMAC.

On a per-learner perspective, the picture is diverse. Differential Evolution clearly outperforms SMAC for Random Forest and AdaBoost. For both SVM algorithms, SMAC wins on more than half of the datasets - even after tie-breaking. For kNN, both tuning methods achieve equal performance after breaking the ties. For MLP there is only a single tie and Differential Evolution wins in the majority of cases.

\begin{table*}
\centering
\begin{scriptsize}
\begin{tabular}{ r | c | c | c | c | c | c | l  }
	%
				& kNN  			& Linear  		& Kernel 	& AdaBoost& Random 		& MLP			&  sum 								\\ 
				& 					& SVM					& SVM			&					&	Forest		& 				& 										\\\hline

	DE		& 7	 (25)		& 6 (18)			& 16 (20)	& 38 (40)	& 34 (39)  	& 28 (28)		& \textbf{129 (170)} \\
	SMAC	& 17 (24)		& 27 (31)			& 28 (29)	& 8 (9)		& 8 (10)		& 20 (21)		& \textbf{108 (124)} \\
	tie 	& 25 (0)		& 16 (0)			& 5	 (0)	& 3	(0)		& 7 (0)			& 1 (0)			& \textbf{57 (0)}  \\

\end{tabular}
\end{scriptsize}
\vspace{.1cm}
\caption{Experiment 1 with a budget of 1 hour per experiment run per dataset. The results are based on five-fold cross-validation mean balanced error. Numbers in brackets apply after ties are broken based on first to reach the reported minimum error. Aggregate results bold for better readability.\label{tab:experiment1}}
\end{table*}

\subsection{Experiment 2}
\subsubsection{Relative Performances}

\begin{table}
\centering
\begin{adjustbox}{angle=90}
\begin{scriptsize}
\begin{tabular}{ r | c | c | c | c | c | c | c | c | c | c | c | c | c | c | c | c | c | c | c | c | l  }
	 & \multicolumn{3}{|c|}{kNN} & \multicolumn{3}{|c|}{Linear SVM} &  \multicolumn{3}{|c|}{Kernel SVM}&  \multicolumn{3}{|c|}{AdaBoost} & \multicolumn{3}{|c|}{Random Forest}  & \multicolumn{3}{|c|}{MLP}&  \multicolumn{3}{|c}{sum}\\ 
 ID 	& DE 			& tie 	& SMAC 				& DE & tie & SMAC 	& DE & tie & SMAC & DE & tie & SMAC & DE & tie & SMAC & DE & tie & SMAC & DE & tie & SMAC \\ \hline	
46& 1(4) & 4(0) & 0(1)& 0(0) & 2(0) & 3(5)& 1(1) & 0(0) & 4(4)& 5(5) & 0(0) & 0(0)& 4(5) & 1(0) & 0(0)& 5(5) & 0(0) & 0(0)& 16(20) & 7(0) & 7(10) \\
184& 4(4) & 0(0) & 1(1)& 2(2) & 0(0) & 3(3)& 4(4) & 0(0) & 1(1)& 3(3) & 0(0) & 2(2)& 2(4) & 2(0) & 1(1)& 3(3) & 0(0) & 2(2)& 18(20) & 2(0) & 10(10) \\
293& 0(0) & 5(5) & 0(0)& 1(1) & 0(0) & 4(4)& 3(3) & 2(2) & 0(0)& 0(0) & 0(0) & 5(5)& 0(0) & 0(0) & 5(5)& 3(3) & 0(0) & 2(2)& 7(7) & 7(7) & 16(16) \\
389& 0(0) & 2(0) & 3(5)& 2(2) & 0(0) & 3(3)& 1(1) & 0(0) & 4(4)& 5(5) & 0(0) & 0(0)& 0(0) & 0(0) & 5(5)& 1(1) & 0(0) & 4(4)& 9(9) & 2(0) & 19(21) \\
554& 0(0) & 4(4) & 1(1)& 5(5) & 0(0) & 0(0)& 1(1) & 4(4) & 0(0)& 2(2) & 0(0) & 3(3)& 2(2) & 0(0) & 3(3)& 0(0) & 0(0) & 5(5)& 10(10) & 8(8) & 12(12) \\
772& 1(1) & 0(0) & 4(4)& 0(5) & 5(0) & 0(0)& 5(5) & 0(0) & 0(0)& 4(4) & 0(0) & 1(1)& 4(5) & 1(0) & 0(0)& 1(1) & 0(0) & 4(4)& 15(21) & 6(0) & 9(9) \\
917& 0(4) & 4(0) & 1(1)& 2(5) & 3(0) & 0(0)& 1(1) & 0(0) & 4(4)& 3(3) & 0(0) & 2(2)& 5(5) & 0(0) & 0(0)& 3(3) & 0(0) & 2(2)& 14(21) & 7(0) & 9(9) \\
1049& 0(0) & 3(0) & 2(5)& 5(5) & 0(0) & 0(0)& 4(4) & 0(0) & 1(1)& 4(4) & 0(0) & 1(1)& 4(4) & 0(0) & 1(1)& 2(2) & 0(0) & 3(3)& 19(19) & 3(0) & 8(11) \\
1120& 4(4) & 1(0) & 0(1)& 0(0) & 0(0) & 5(5)& 1(1) & 0(0) & 4(4)& 4(4) & 0(0) & 1(1)& 2(2) & 0(0) & 3(3)& 0(0) & 0(0) & 5(5)& 11(11) & 1(0) & 18(19) \\
1128& 2(3) & 1(0) & 2(2)& 0(2) & 4(0) & 1(3)& 3(4) & 1(0) & 1(1)& 4(4) & 0(0) & 1(1)& 3(3) & 0(0) & 2(2)& 3(3) & 0(0) & 2(2)& 15(19) & 6(0) & 9(11) \\ \hline
sum& 12(20) & 24(9) & 14(21)& 17(27) & 14(0) & 19(23)& 24(25) & 7(6) & 19(19)& 34(34) & 0(0) & 16(16)& 26(30) & 4(0) & 20(20)& 21(21) & 0(0) & 29(29) & \textbf{134(157)} & \textbf{49(15)} & \textbf{117(128)}\\
\end{tabular}
\end{scriptsize}
\end{adjustbox}
\vspace{.1cm}
\caption{Experiment 2 with a budget of 12 hours. Statistics of five experiment runs for six classifiers and ten datasets per tuning method. Aggregate results bold for better readability.\label{tab:budget_100}}
\end{table}

Table \ref{tab:budget_100} lists the counts of experiment runs that Differential Evolution wins or ties against SMAC. When aggregated over all learners and datasets (bold entries), Differential Evolution outperforms SMAC by scoring 14.5\% more wins. Similar to Experiment 1, breaking the ties benefits Differential Evolution: after tie-breaking, it scores 22.7\% more wins than SMAC.


\begin{figure}[t!]
\centering
\includegraphics[width=0.3\textwidth]{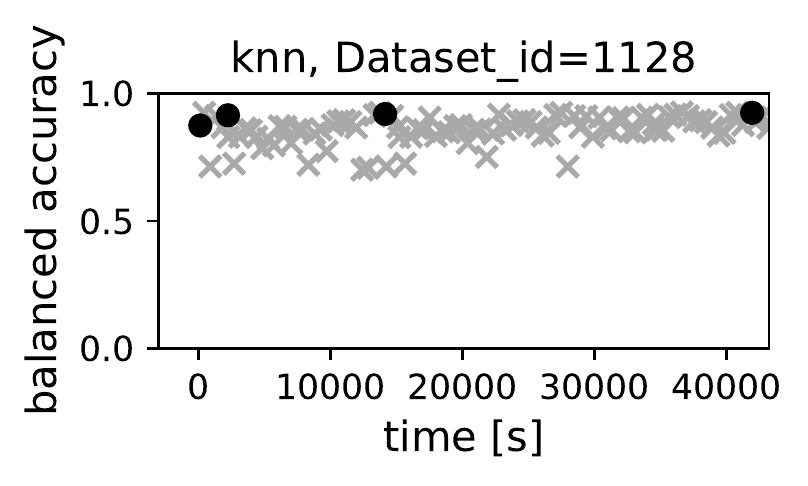}
\caption{Hyperparameter tuning progress of Differential Evolution (gray, cross) and SMAC (black, dot) of kNN on dataset 1128. Result: tie (to be broken in favor of Differential Evolution). \label{fig:knn_learning}}
\end{figure}

%

On a per-learner perspective, i.e., aggregated over all datasets, Table \ref{tab:budget_100} indicates that for kNN, and linear SVM both methods perform similarly, even after breaking the ties. Kernel SVM is slightly favorable to Differential Evolution with 26.3\% more wins than SMAC before tie-breaking, 31.6\% more wins after tie-breaking. For Random Forest, Differential Evolution scores significantly more wins (30\% more before tie-breaking, 50\% after), while for MLP SMAC does (38.1\% more wins than Differential Evolution, no ties). Differential Evolution most clearly outperforms SMAC for AdaBoost by achieving more than twice the number of wins of SMAC. When excluding AdaBoost from the experiment results, Differential Evolution (100 wins) is on par with SMAC (101 wins) before tie-breaking. After tie-breaking, Differential Evolution scores 9.8\% more wins than SMAC (123 wins to 112). 

\begin{figure}[t!]
\centering
\includegraphics[width=0.3\textwidth]{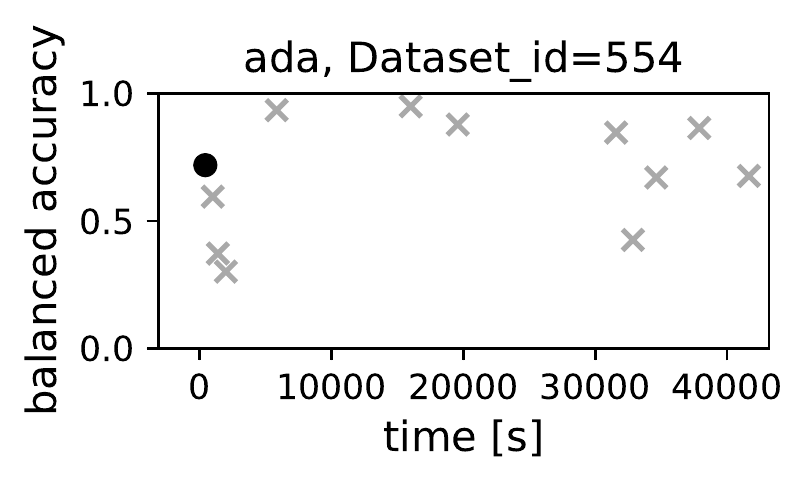}
\caption{Hyperparameter tuning progress of Differential Evolution (gray, cross) and SMAC (black, dot) of AdaBoost on dataset 554. Result: Differential Evolution wins.\label{fig:ada_learning554}}
\end{figure}

%

On a per-dataset perspective, Table \ref{tab:budget_100} indicates that Differential Evolution performs as good as or better than SMAC for most datasets. Before as well as after tie-breaking, SMAC scores more wins on datasets 293, 389, 554, and 1120. Conversely,  Differential Evolution achieves a lower mean balanced error in cross-validation than SMAC on six of the ten datasets. Even when ignoring AdaBoost in the results, the base learner that Differential Evolution most clearly outperformed SMAC on, 
SMAC still only wins four out of ten datasets.

\subsubsection{Learning Curves - Progress of Hyperparameter Tuning}


Figure~\ref{fig:knn_learning} and~\ref{fig:ada_learning554} 
visualize balanced accuracy over time for selected learners and datasets. Crosses represent tested individuals in the population of Differential Evolution. 
The prominent horizontal spacing of crosses indicates the time needed to complete the training of base learners with a hyperparameter configuration. The figures also illustrate SMAC's progress of tuning hyperparameters. As we used a SMAC implementation logging only timing information when identifying a new best configuration, the figures do not provide information about the training time needed for different hyperparameter configurations chosen by SMAC between two best configurations. 

Figure \ref{fig:knn_learning} 
shows steady learning progress for both hyperparameter tuners on dataset 1128 
for the kNN classifier. 
The base learner can process the dataset quickly; therefore no horizontal white spaces are observable for the Differential Evolution plot. The figure depicts a tie between the tuners. 

Figure \ref{fig:ada_learning554} visualizes a case when Differential Evolution outperforms SMAC when tuning AdaBoost for dataset 554. The plot exhibits prominent horizontal spacing for Differential Evolution. The few plotted crosses show that not even the initial population could be evaluated entirely. That means evolution did not start before the budget ran out - a problem for large datasets. 

\begin{figure}[t!]
\centering
\includegraphics[width=0.4\textwidth]{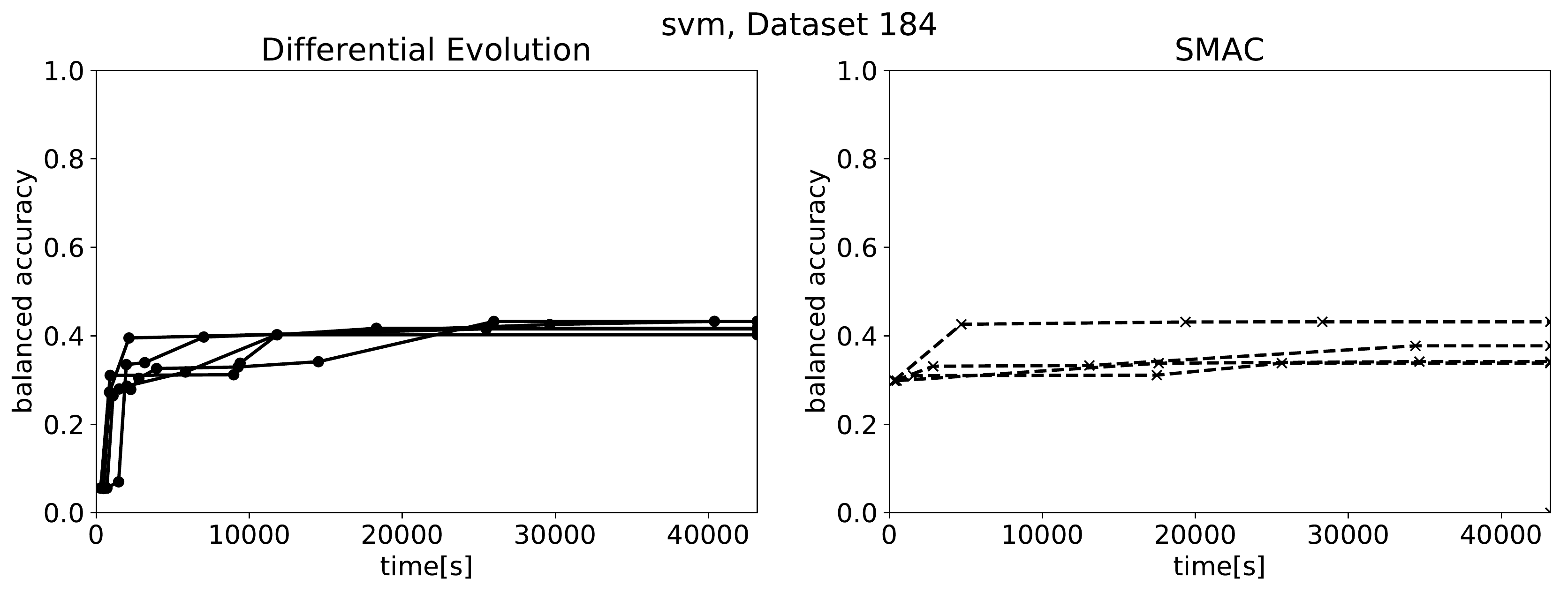}
\caption{Differential Evolution (left) and SMAC (right) learning curves for tuning kernel SVM hyperparameters on dataset 184.\label{fig:svm_lc184}}
\end{figure}
\begin{figure}[t!]
\centering
\includegraphics[width=0.4\textwidth]{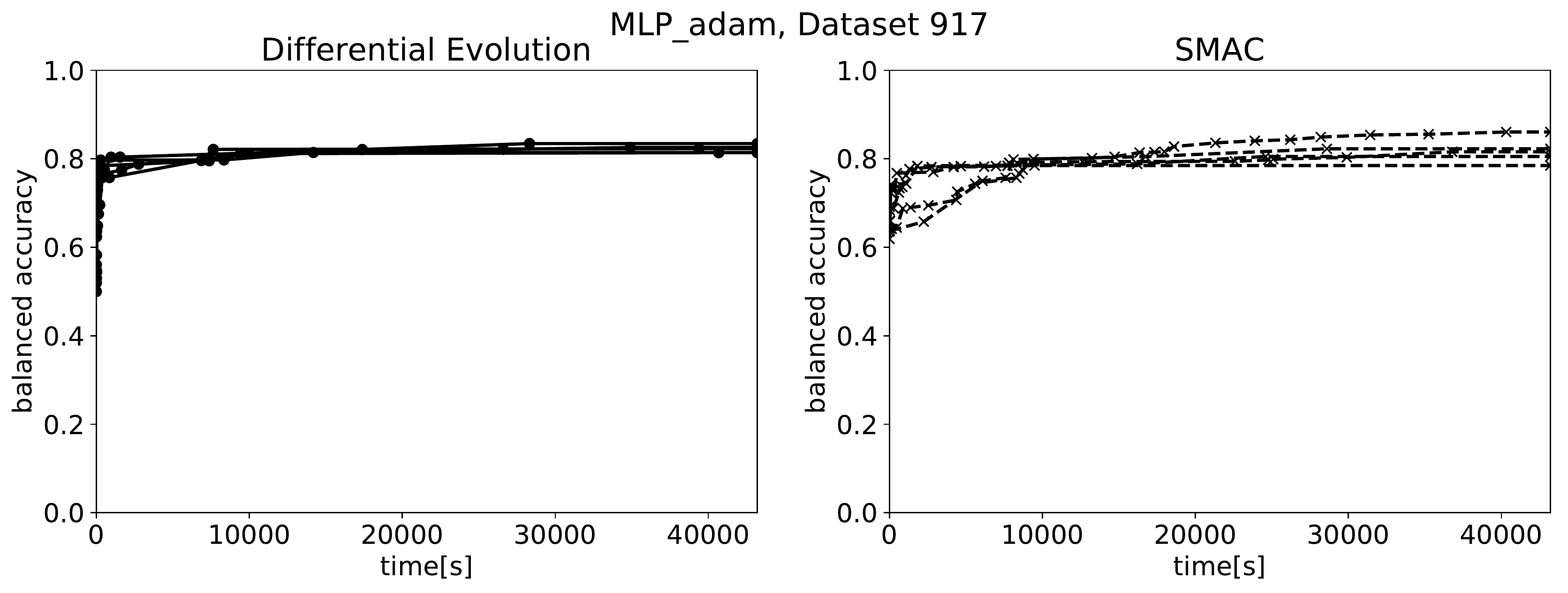}
\caption{Differential Evolution (left) and SMAC (right) learning curves for tuning MLP hyperparameters on dataset 917.\label{fig:mlp_lc917}}
\end{figure}
\begin{figure}[t!]
\centering
\includegraphics[width=0.4\textwidth]{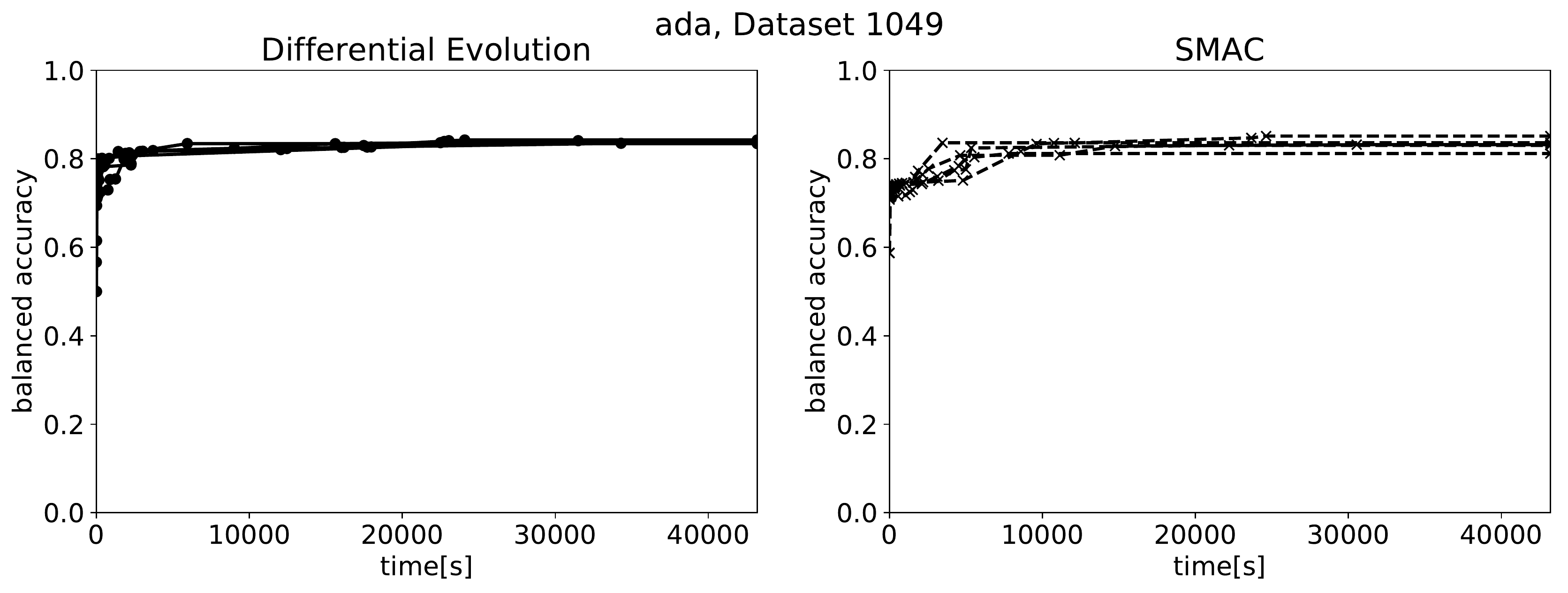}
\caption{Differential Evolution (left) and SMAC (right) learning curves for tuning AdaBoost hyperparameters on dataset 1049.\label{fig:ada_lc1049}}
\end{figure}

Figure \ref{fig:svm_lc184}, \ref{fig:mlp_lc917}, and \ref{fig:ada_lc1049} illustrate the maximum reported balanced accuracy over time for each experiment run repetition for both tuning methods. The curves indicate consistent learning behavior. Note that for other base learners and datasets (not shown), the recorded learning curves vary even less.

\section{Discussion}
\subsection{Experiment Results}
\textbf{Experiment run statistics.} Typical evolutionary algorithms do not rely on a model of the process to be optimized but rather on random chance and the algorithmic equivalents of biological evolution. When compared to model-based methods such as Bayesian Optimization in the context of hyperparameter tuning, this may or may not represent a drawback. As typical evolutionary methods do not use gradients in their optimization progress, they usually have to repeat objective function evaluation more often than gradient-based methods. However, Bayesian Optimization might be misled if its model $p_i(f | \mathcal{D})$ should be ill-suited for the specific base learner whose hyperparameters are tuned. In this respect, the experiments show interesting results: Differential Evolution performs at least as good as SMAC, a Bayesian Optimization method leveraged in auto-sklearn \cite{feurer2015efficient}, for a variety of datasets and across a range of machine learning algorithms. In fact, Differential Evolution scores more wins than SMAC both in Experiment~1 (19.4\% more wins without tie-breaking, 37.1\% with tie-breaking) and~2 (14.5\%, 22.7\%). 

Figure \ref{fig:svm_lc184} - \ref{fig:ada_lc1049} exhibit consistent behavior for each tuning method, which we also confirmed for other base learners and datasets (not shown). That suggests that experiment runs per tuning method, base learner, and dataset are sufficiently informative for analyzing Experiment 2 results.

For Experiment 2 MLP is the only base learner on which SMAC significantly outperforms Differential Evolution. Both tuning algorithms perform similarly on two learners (kNN and linear SVM), and Differential Evolution outperforms SMAC on three learners (kernel SVM, Random Forest, AdaBoost) with the most definite results for AdaBoost. According to the experimental results of \cite{feurer2015efficient}, AdaBoost performs well compared to other learning algorithms on most datasets. Thus, Differential Evolution's strong performance in both experiments for AdaBoost suggests to use it rather than SMAC for tuning AdaBoost's hyperparameters. Note that the per-learner tendencies between Experiment 1 and Experiment 2 differ for kNN, linear SVM, and kernel SVM: without tie-breaking SMAC wins more often in Experiment 1, but not so in Experiment 2. Also for MLP, the results reverse between both experiments: Differential Evolution wins more often in Experiment 1, but SMAC in Experiment 2. Only AdaBoost and Random Forest are winners for Differential Evolution in Table \ref{tab:experiment1} and \ref{tab:budget_100}.  

Tie-breaking usually favors Differential Evolution, i.e., it is faster to report the maximum accuracy achieved by both tuning methods. SMAC outperforms Differential Evolution in the early stages - in particular on the bigger datasets, if the budget is too short for the evolution phase to make significant tuning progress or even to start at all.

Table \ref{tab:budget_100} states that tie-breaking does not resolve 15 ties for datasets 293 and 554. This only occurs if a given SMAC experiment run and its Differential Evolution counterpart do not report a single evaluation result within the time limit. That indicates that the 12 hour time budget is challenging for these datasets, in particular when tuning the hyperparameters of kNN and kernel SVM (Table \ref{tab:budget_100}). In Experiment 2, datasets 293 and 554 are the largest (number of samples times the number of features per sample, see \cite{feurer2015efficient}). Figure \ref{fig:ada_learning554} illustrates the learning progress of AdaBoost on dataset 554. The prominent horizontal spacing for Differential Evolution's learning curve confirms that large datasets require substantial computation time to train and test a single hyperparameter configuration - note the difference to Figure \ref{fig:knn_learning} on the smaller dataset 1128.

\begin{figure}[t!]
\centering
\includegraphics[width=0.35\textwidth]{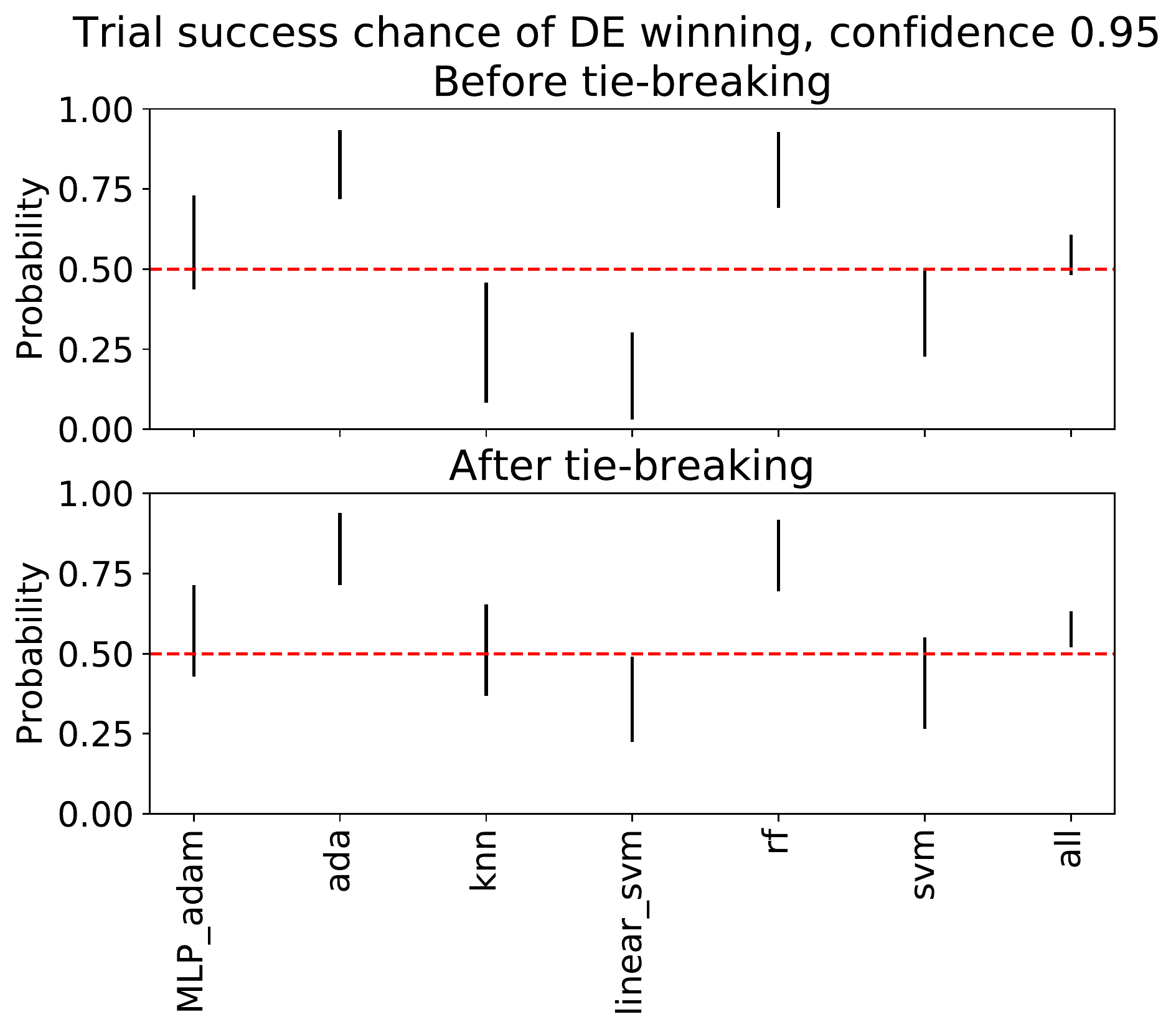}
\caption{Experiment 1: Confidence Intervals of 95\% confidence for the chances of Differential Evolution outperforming SMAC - per base learner. For reference, the red dashed line indicates 50\% chances of success. 'all' corresponds to the sum column of Table \ref{tab:experiment1}. \label{fig:perAlgSmall-CI}}
\end{figure}

\begin{figure}[t!]
\centering
\includegraphics[width=0.35\textwidth]{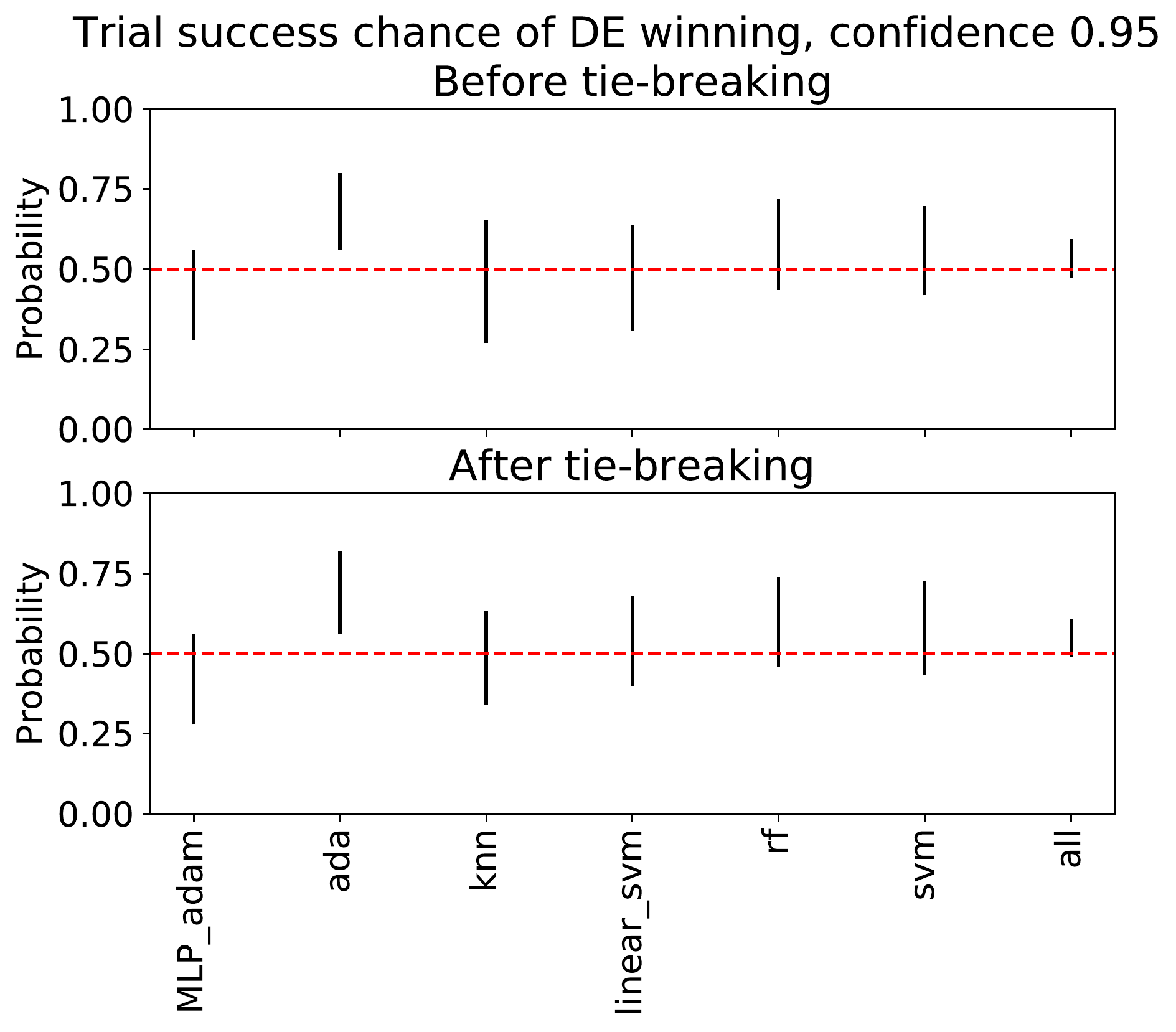}
\caption{Experiment 2: Confidence Intervals of 95\% confidence for the chances of Differential Evolution outperforming SMAC - per base learner. For reference, the red dashed line indicates 50\% chances of success. 'all' corresponds to the bold sum entries in Table \ref{tab:budget_100}. \label{fig:perAlg-CI}}
\end{figure}

\begin{figure}[t!]
\centering
\includegraphics[width=0.35\textwidth]{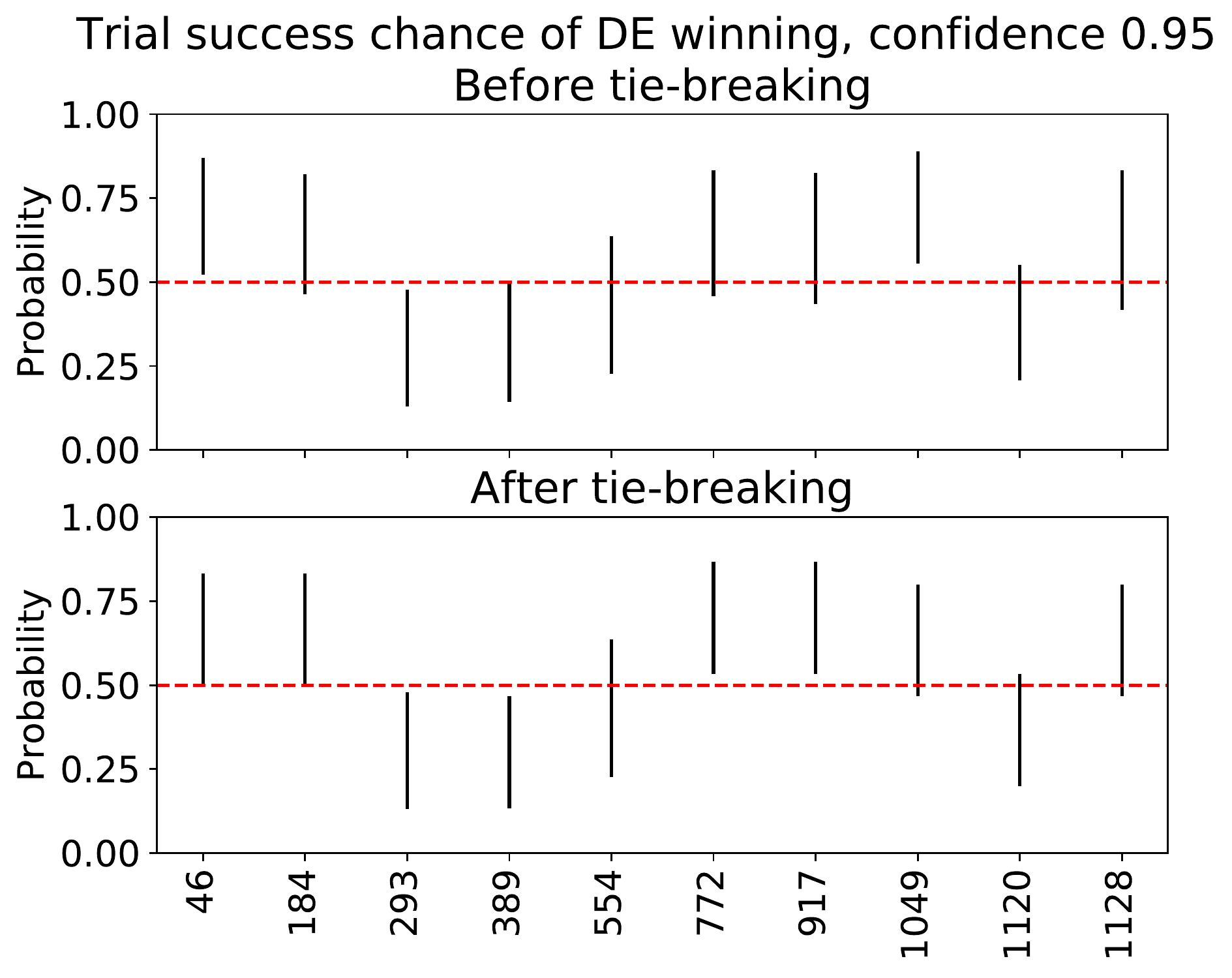}
\caption{Experiment 2: Confidence Intervals of 95\% confidence for the chances of Differential Evolution outperforming SMAC - per dataset. For reference, the red dashed line indicates 50\% chances of success. \label{fig:perDS-CI}}
\end{figure}

\textbf{Inferential statistics.} Figure \ref{fig:perAlgSmall-CI}, \ref{fig:perAlg-CI}, and \ref{fig:perDS-CI} illustrate the intervals of 95\% confidence of the Bernoulli trial probability of Differential Evolution successfully outperforming SMAC. The breaking of ties generally favors Differential Evolution, and there is a noticeable upward shift for many of the confidence intervals after tie-breaking. Most per-algorithm and per-dataset confidence intervals in Figures \ref{fig:perAlgSmall-CI}-\ref{fig:perDS-CI} cross the 50\% reference line, suggesting that there is no significance at the 95\% level for a success probability strictly above or below 50\%. That implies that these confidence intervals do not provide statistical justification to prefer either tuning method. However, several of these confidence intervals tend to favor Differential Evolution  - larger shares reside above the reference line than below. With additional experiment runs in the future, the confidence intervals should shrink, and success or failure probability may become statistically significant. Figure \ref{fig:perDS-CI} shows that with high confidence, the Differential Evolution results are negative for datasets 293 and 389 as the intervals' upper bounds stay below the 50\% reference line. Also, dataset 1120, and to a much lesser extent also dataset 554, tends to favor SMAC as a larger portion of the confidence interval resides below the reference line. When aggregating for Experiment 1 all base learners after tie-breaking, Figure \ref{fig:perAlgSmall-CI} suggests a statistically significant result in favor of Differential Evolution. After tie-breaking, Figures \ref{fig:perAlgSmall-CI} and \ref{fig:perAlg-CI} suggest statistical significance of Differential Evolution outperforming SMAC for tuning AdaBoost in both experiments. For Random Forest results also tend to favor Differential Evolution, however less strongly. It is striking that both ensemble-based methods (AdaBoost, Random Forest) are favorable to Differential Evolution and the results are less clear or negative for the other learning algorithms. However, we have not been able to identify the algorithmic reason for this behavior yet, and it remains a research question to investigate what determines the tuners' performance when tuning different learners' hyperparameters, and why. 

For the total aggregate of Experiment 2, Figure \ref{fig:perAlg-CI} shows that even after tie-breaking the confidence interval crosses the reference line - its lower bound reaches 49\%. Statistical t-tests confirm that Differential Evolution's total aggregate success chance being larger than 50\% in Experiment 2 is not significant at the 95\% confidence level (but it is at the 90\% level, results omitted for brevity). However, it is close to being statistically significant at the 95\% level. Overall, the statistical results are encouraging future work. We anticipate that several of the advanced  Differential Evolution variants in \cite{5601760}\cite{ALDABBAGH2018284} will improve on our experiment results and tip the scale against Bayesian Optimization.

By design, the experimental setup is meant to present a challenge for both hyperparameter tuning methods to investigate their tuning performance relative to one another. In this setup, both tuning methods may suffer from large datasets, limited CPU resources, and tight time budgets. Note that here the iterative approach of Bayesian Optimization is a strength when compared to Differential Evolution. As Bayesian Optimization collects new samples, it updates its probabilistic model. Even if the time budget is small, as long as it evaluates more than a single hyperparameter configuration, the successive iterations should sample better and better configurations. On the other hand, if Differential Evolution evaluates the same number of hyperparameter configurations, as long as the number of evaluations is smaller than or equal the population size, its evolution has not started, yet. In that situation, no improvements of the hyperparameter configurations are to be expected, and performance is a matter of chance. A possible way to improve its relative performance on large datasets such as 293 or 554 for which it does not finish evaluating the initial population could be to reorder the initial population members for increasing (expected) computational cost. This way, Differential Evolution can evaluate at least more configuration-dataset samples within the budget. In addition, reducing the population size by shrinking $n$ when facing very tight time budgets could help Differential Evolution reach the evolution operations earlier. However, that reduces the exploration potential of the method.

\begin{figure}[t!]
\centering
\includegraphics[width=0.35\textwidth]{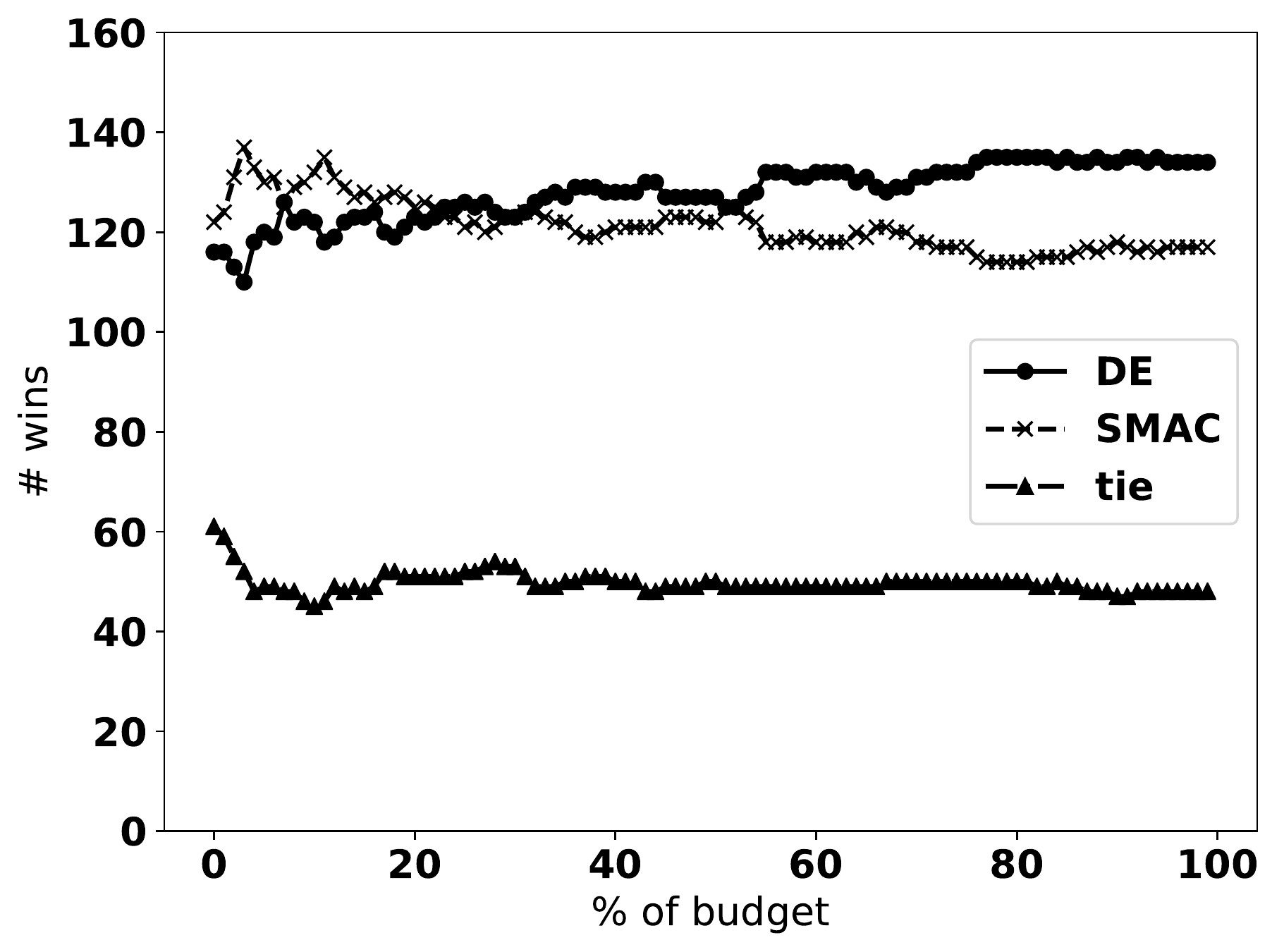}\caption{Experiment 2 results without tie-breaking for different time budgets in 1\% increments (100\%=12 hours).\label{fig:budget breakdown results}}
\end{figure}

\begin{figure}[t!]
\centering
\includegraphics[width=0.35\textwidth]{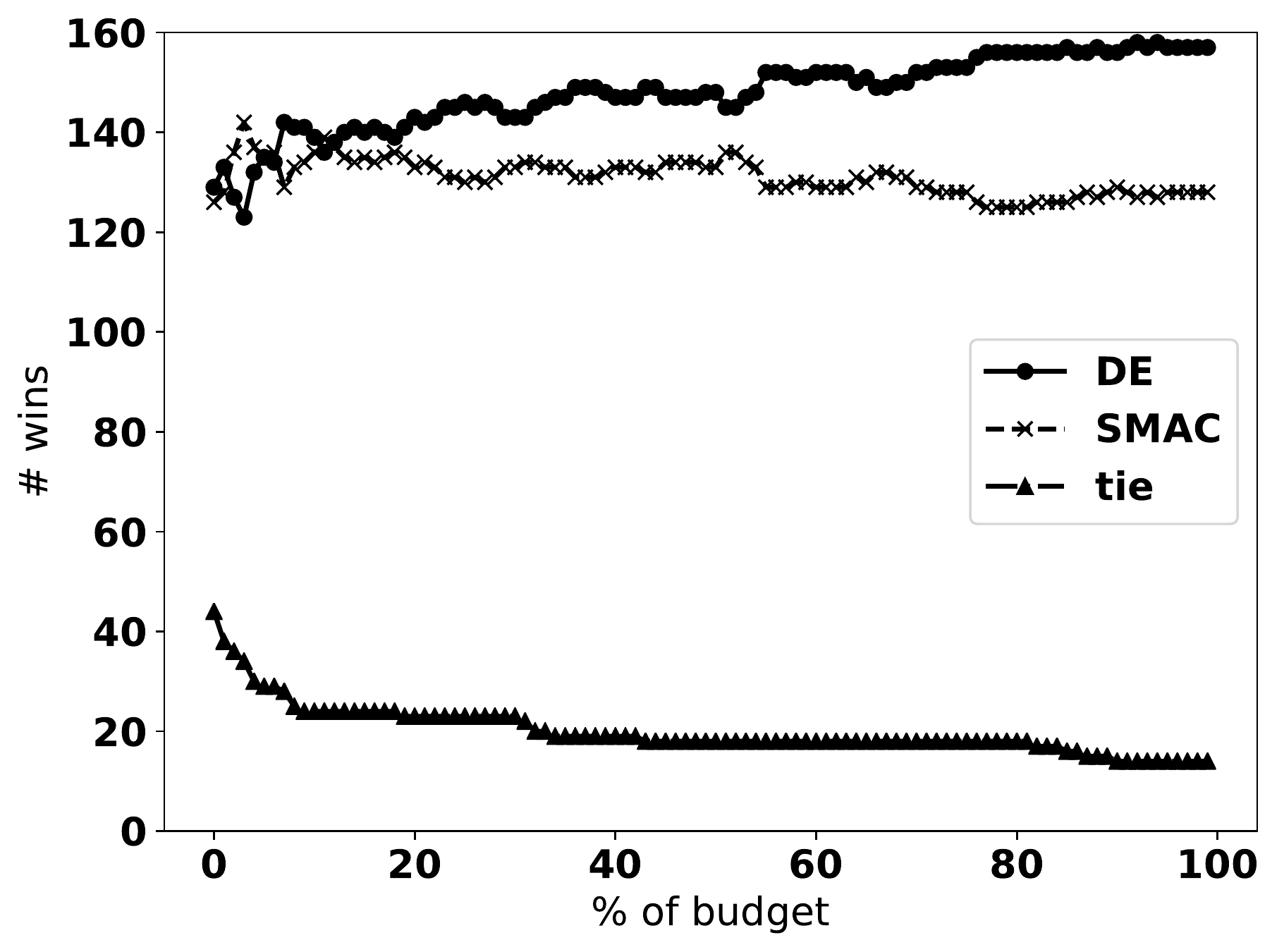}\caption{Experiment 2 results as in Figure \ref{fig:budget breakdown results}, but with tie-breaking.\label{fig:budget breakdown results tiebreak}}
\end{figure}

\textbf{Tighter time budgets.} 
Figure \ref{fig:budget breakdown results} and \ref{fig:budget breakdown results tiebreak} illustrate the Experiment 2 results had different shares of the original 12-hour budget been applied. As more computing resources become available, Differential Evolution improves in performance relative to SMAC. At a budget of approximately 30\% (4 hours), it crosses SMAC's score and consistently remains above it. The figures confirm that breaking ties is favorable to Differential Evolution. With tie-breaking, it scores more wins than SMAC for even smaller budgets. A budget of less than 10\% (1.2 hours) is required for Differential Evolution to first score more wins than SMAC, with a short period of reversal at a budget of 15\%. For larger budgets, Differential Evolution consistently achieves more wins than SMAC. The gap widens as the budget increases.   

\subsection{Limitations}

This work assesses the suitability of the selected optimization approaches for hyperparameter tuning. Therefore it focuses exclusively on cold start situations and does not consider other relevant aspects such as meta-learning, ensembling, and data preprocessing steps. Future work will extend to these.  

The experimental setup limits the execution of experiment runs to a single CPU core. That reduces the potential impact of how well the used software libraries and frameworks can exploit parallelism. The achievable performance gains also depend on the base learner's capability of using parallel computing resources. For example, Random Forest is an ensemble method parallel in the number of trees used, whereas AdaBoost is sequential due to the nature of boosting. Future work will study the impact of parallelism on the hyperparameter tuning performance for different methods and base learners. 



\section{Conclusion and Future Work}
This paper compares Differential Evolution (a well-known representative of evolutionary algorithms) and SMAC (Bayesian Optimization) for tuning the hyperparameters of six selected base learners. 
In two experiments with limited computational resources (single CPU core, strict wall-clock time budgets), Differential Evolution outperforms SMAC when considering final balanced classification error. In Experiment~1, the optimization algorithms tune the hyperparameters of the base learners when applied to 49 different small datasets for one hour each. In Experiment 2, both optimization algorithms tune the base learners’ hyperparameters for 12 hours each when applied to ten different representative datasets. In the former experiment, Differential Evolution scores 19\% more wins than SMAC, in the latter 15\%. The results also show that Differential Evolution benefits from breaking ties in a `first-to-report-best-final-result' fashion: for Experiment 1, Differential Evolution's wins 37\% more often than SMAC, in Experiment 2 23\%. Experiment 2 also shows that only when the budget is tiny, SMAC performs better than Differential Evolution. That occurs when Differential Evolution is late to enter the evolution phase or not even able to finish evaluating the initial population. Differential Evolution is particularly strong when tuning the AdaBoost algorithm. Already with the basic version of Differential Evolution, positive results can be reported with statistical significance for some of the datasets and base learners. That suggests considerable potential for improvements when using some of the improved versions in \cite{5601760}\cite{ALDABBAGH2018284}.

We see several possibilities to extend this work. First, future work should study if more recent evolutionary algorithms such as the variants of Differential Evolution listed in \cite{5601760}\cite{ALDABBAGH2018284} can improve hyperparameter tuning results. 
A second avenue is to integrate meta-learning \cite{AAAI1817235} by choosing the initial population's parameters accordingly. Then,  Probabilistic Matrix Factorization approaches such as \cite{NIPS2018_7595} will also have to be considered for comparison. 
Third, an investigation is required to understand why Differential Evolution performs better than SMAC when tuning some of the base learners, while the results are less clear or negative for the other learning algorithms. Fourth, we intend to investigate whether hybrid methods as \cite{falkner2018bohb} could benefit from adopting concepts of evolutionary algorithms. Finally, future work will extend to entire machine learning pipelines, i.e., to support also preprocessing steps and ensembling as \cite{feurer2015efficient} and study the implications of parallel execution.


\bibliography{automl}
\bibliographystyle{IEEEtran} 
\end{document}